\documentclass[11pt]{article}

\usepackage{amsmath,amssymb,amsfonts}
\usepackage{algorithmic}
\usepackage{graphicx}
\usepackage{textcomp}
\usepackage{hyperref}
\usepackage{cleveref}
\usepackage[numbers]{natbib}
\usepackage{caption}
\usepackage{soul,color}
\usepackage{multirow}
\usepackage{ctable}
\usepackage{stfloats}
\usepackage{xcolor}
\usepackage{siunitx}
\usepackage{booktabs}
\usepackage{listings}
\usepackage{enumitem}
\usepackage{xspace}
\usepackage{courier}

\newcommand{\algname}[0]{\texttt{GAversary}\xspace}

\newcommand{\added}[1]{#1}
\newcommand{\deleted}[1]{}

\def\BibTeX{{\rm B\kern-.05em{\sc i\kern-.025em b}\kern-.08em
    T\kern-.1667em\lower.7ex\hbox{E}\kern-.125emX}}
\begin{document}

\title{Vulnerability of Natural Language Classifiers to
Evolutionary Generated Adversarial Text}

\author{
Manjinder Singh\\
Computing Science \& Mathematics\\
University of Stirling, Stirling, UK
\and
Alexander E. I. Brownlee\\
Computing Science \& Mathematics\\
University of Stirling, Stirling, UK\\
\texttt{alexander.brownlee@stir.ac.uk}
\and
Mohamed Elawady\\
Computer and Information Sciences\\
University of Strathclyde, Glasgow, UK
}

\date{}

\maketitle

\begin{abstract}
Deep learning models have achieved impressive performance across various fields but remain vulnerable to adversarial inputs, particularly in NLP, where such attacks can have significant real-world consequences. Adversarial attacks often involve small, semantically similar token replacements to fool NLP models, and recent methods have become more precise by targeting specific vulnerable words, often by exploiting some level of access to the model's internal structure. This paper proposes \algname, a hybrid Genetic Algorithm (GA) to generate adversarial attacks on natural language models. The GA is able to treat the target model as a black box, requiring only the logit value output by the model to guide the search. \algname differs from GAs previously proposed for this problem, by using GloVe embeddings to propose word replacements (the mutation operator) to improve the semantic similarity of the adversarial examples. \algname is applied to several benchmark data sets and well-known target models. \algname is able to substantially reduce the target model's accuracy on test data compared to the BAE and A2T attacks compared against (in the best case, reducing a 76.8\% accuracy to 5.8\%, compared to BAE's 27.6\%). The trade-off is that \algname perturbs just under twice as many words as the other two methods, with a slightly lower semantic similarity to the original text, with around a 5\% increase in run-time.
\end{abstract}


\section{Introduction} 
\label{sec:introduction}
Deep learning models such as deep neural networks (DNNs) have been applied to a vast range of fields including speech recognition, natural language processing (NLP) and computer vision. In many areas they have performed as well if not better than human experts \citep{gebru2017using}. However, they are vulnerable to adversarial inputs; perturbations that may be imperceptible to humans but can cause DNNs to misclassify the output \citep{dodge2017study}.

Automated text analysis through NLP models is increasingly used to auto-generate summaries which drive, for example, news recommender websites, with direct implications for reputation, reliability, and branding \citep{reuver-etal-2021-nlp}. For example, NLP models can be used to inform investors on buy/sell decisions based on the prevailing sentiment within these reviews \citep{angeles2020investor,lucey2005role}. Thus, adversarial attacks on NLP models could have major impact, such as triggering poor investment decisions and the corresponding disruption of local economies.

The overall goal for an adversarial system is to generate an attack, in the form of changes to an original text such that the original text is misclassified, having the least number of modifications to the original text, and maintaining semantic similarity after those modifications.
A variety of approaches have been demonstrated in the literature. Generally these rely on replacing tokens in the examples \citep{roth2021token}.
Typically such approaches rely on generating a few synonym token replacements (usually using some form of semantic similarity), and choosing the one replacement that has the biggest influence on the classification probability for the targeted model \citep{li2018textbugger,gao2018black}.
Recently these approaches have become more targeted, identifying the most vulnerable words, and using context to select suitable replacements \citep{li2020bert,li2020contextualized}.
However, the search process each of these methods follows generally amounts to a variation of local or some other single-point search.
Genetic Algorithms (GAs) perform well on combinatorial optimization problems where \deleted{we have}\added{there is} a large search space \citep{anderson1994genetic,bennis2020nature}. Even with only the combination of synonyms (ignoring sentence structure), Natural Language has such a vast space to explore, yet there are few examples of the application of GAs to generating adversarial examples.

The advantage of GAs is that they can treat the target problem as a black box, which makes them particularly suitable for an attack scenario in which the inner workings of the targeted model are unavailable.
As GAs follow a search process that simply seeks inferences from the target model and can be terminated at any arbitrary point, they also have the potential to generate attacks more quickly than GANs.
Thus, \deleted{in the present work we present}\added{this paper presents} \textit{\algname}, which uses a Genetic Algorithm to target several well-known pre-trained models.
The other recent attempts to use GAs to target NLP models rely on operators that work at random and do not exploit the targeting and context awareness of Bert-Attack \citep{li2020bert,li2020contextualized} and others. More recently \cite{ye2021heuristic}, GloVe has been used to guide the search as \deleted{we do}\added{\algname does}, but they limit the search to a filtered set of `influential' words, whereas \deleted{we allow}\added{\algname allows} the search to explore a wider range of possible replacements. Another approach \cite{alzantot2018generating} also used GloVe, and targeted a wider range of words for replacement, but used GloVe to identify semantically similar words as replacements, rather than \deleted{our approach}\added{\algname} which uses the surrounding context to identify replacements. \deleted{Our}\added{The paper's} major contribution then is to draw on the GloVe embedding model to guide the genetic operators towards contextually-plausible equivalent word replacements. 
In this setting, the GA still treats the target model as a black box; the use of GloVe is independent of whatever model is being targeted.

\deleted{Our}\added{This paper's} research questions are as follows:

\begin{enumerate}
    \item Can a genetic algorithm guided by GloVe embeddings generate adversarial examples for a pre-existing NLP model such that the model's prediction accuracy is significantly reduced for those examples?

    \item What is the performance of the proposed algorithm compared to existing methods, in both reducing accuracy of the target model, and computational efficiency?

    \item How different do the examples need to be from the original text for the model to misclassify them for \deleted{our method}\added{\algname}?
\end{enumerate}

\deleted{We perform thorough experimentation comparing our approach}\added{Thorough experimentation compares \algname} to many models, alternative techniques, and data sets, to answer these questions. \deleted{Our proposed approach}\added{\algname} seeks the best of previous work: like Bert-Attack it is fast as it only needs to run target model, and uses GloVe for semantic similarity. However, while \deleted{we also use}\added{\algname also uses} nothing but the logit value from the target model (the same as Bert-Attack), that approach ranks words by sensitivity, then iterates over the ranking making substitutions exhaustively. In contrast, \deleted{we pick}\added{\algname picks} words at random to avoid local optima, and guides the overall search using a genetic algorithm. Previous GA-based approaches either do not use the guidance from the target model in the way \deleted{we do}\added{\algname does}, or do not include the context of words being replaced. Thus our key novelty is proposing a hybrid approach getting the best of both Bert-Attack's guided mutations to the target text, and the Genetic Algorithm's high level efficient search. \added{The main contributions of this work, then, are: (1) \algname, a hybrid approach to generating adversarial text using a genetic algorithm search guided by GloVe for semantic similarity, and (2) experimental results showing that the approach reduces accuracy more than other state-of-the-art techniques.}



\added{The paper continues in \Cref{sec:related} by reviewing related work in adversarial attacks, with a focus on natural language. \Cref{sec:method} introduces \algname, including the major components of its genetic algorithm framework. \Cref{sec:exp} describes the computational experiments and discusses the results, then \Cref{sec:conclusion} concludes the paper and discusses future work.}

\section{Related Work} 
\label{sec:related}




Adversarial attacks were first applied in the field of computer vision to image classifier models, with the term ‘adversarial example’ first introduced in \cite{szegedy2013intriguing}. It was noticed that the researchers could cause the DNN to misclassify images by introducing small perturbations to the images. Application of `noise' to the images led to misclassification by the DNN even when there was no perceptible difference to human observers. As neural networks have increasingly become more popular in a diverse range of fields within the AI space and specifically within the NLP field, the need to ensure the robustness of the models and the predictions they generate has become more urgent. 

The majority of approaches proposed for attacking NLP models focus on replacing tokens in the examples \citep{roth2021token}. These attacks can be applied into different levels: sentence, word, character or a mixture. \deleted{In our work, we focus}\added{The present work focuses} on word-level attack recipes.

TextBugger \cite{li2018textbugger} is the earliest work  which used different levels of sentence modification (i.e. substitution of characters, replacement of words, and insertion of typo letters) to produce adversarial examples. However, Alzantot et al. \cite{alzantot2018generating} developed a genetic algorithm focusing on words replacement in which the mutation operator uses Euclidean distance in the GloVe embedding space to identify semantically similar words as candidate substitution words in the original text. To ensure the correctness of the text semantics, a filter exploiting the Google 1 billion words language model \citep{chelba2013one} is used to exclude words that do not fit in the surrounding context. A fast variant of Alzantot's method \cite{jia2019certified} used different NLP methods ensuring robust word substitutions through Interval Bound Propagation (IBP) \cite{gowal2018effectiveness}. TextFooler \cite{jin2020bert} checked  important words in both semantic and grammatical way before considering these words as possible substitutes until the model’s prediction changes, without understanding how the used text generation method works.

In the applications of BERT \cite{devlin2019bert} (pre-trained masked language model), a method called BERT-Attack \cite{li2020bert} was introduced to identify less important words in input text in order to replace them with acceptable alternatives generated by BERT, satisfying the semantic and grammatical constraints. Another method BAE \cite{garg2020bae} was developed  based on BERT's masked language model (BERT-MLM) to perform text replacements and insertions while considering grammatical and contextual aspects. CLARE \cite{li2020contextualized} used RoBERTa \cite{liu2019roberta} to introduce three perturbation techniques (Replace, Insert, and Merge) using a mask-then-infill procedure, generating grammatically adversarial examples. A2T \cite{yoo2021towards} used DistilBERT \cite{sanh2019distilbert} to build an attack algorithm determining most substitutional candidates based on gradient-based word importance ranking and generating semantically correct adversarial replacements based on counter-fitted word embeddings.

Inspired by input data perturbations in the computer vision domain \cite{Szegedy2013IntriguingPO} to strength-up the loss functions for better image classification, GBDA \cite{guo2021gradient} and TPGD \cite{yuan2023bridge} adapted these perturbations into the NLP domain by investigating the latent space and optimising the gradients of the components of the model used.

In recent years, different methods (i.e. SemAttack \cite{wang2022semattack}, TextHoaxer \cite{ye2022texthoaxer}, TextHacker \cite{yu2022texthacker}, ATGSL \cite{li2023adversarial}) were developed based on the previous work to find most suitable words inside the text for replacement and to substitute these words with candidates while persevering the text semantics and avoiding the grammar errors.


In one of of the closest proposals to \deleted{our work}\added{\algname}, \cite{ye2021heuristic} proposed a genetic algorithm based approach building on the work of \cite{wang2019natural}. In \cite{wang2019natural}, clustering among the embedding vectors was used to identify synonyms for each word. 
The GA selected replacements from these clusters as its mutation operator; the crossover was the same as \deleted{ours}\added{that of \algname} in that it simply took a random combination of the replacements from both parents. \cite{ye2021heuristic} built on this idea to exploit GloVe, as \deleted{we do}\added{\algname does}, to guide the search for replacement words. GloVe embeddings were used to identify the 10 closest synonyms to each word, capped to a maximum Euclidean distance of 0.5 from the original word. From that list, replacement words were selected in descending order of frequency in the dictionary.  However, rather than choosing the words to be replaced as part of the search, they  adopted what they call \textit{word influence} to identify all words to be replaced. This is, essentially, the sensitivity of the target to each word in the input; determined by removing each word and observing the affect on the model output. A part-of-speech constraint was applied to maintain consistency. Their fitness function included cosine similarity to the original text, and their initial GA population was made by already replacing the most influential words with synonyms. The key differences to \deleted{our approach}\added{\algname} are that \deleted{we allow}\added{\algname allows} changes to everything except stopwords, they always change all of the influential words whereas \deleted{our approach}\added{\algname} might ignore some, and \deleted{we place}\added{\algname places} no constraints on part-of-speech/semantic similarity.

\section{Methodology} 
\label{sec:method}
\deleted{Our}\added{The present} work seeks to investigate the creation of adversarial examples against an NLP model. This research area, compared to similar work with image models, is relatively new and so this project will aim to combine work already done in several similar areas \citep{go2009twitter,mrkvsic2016counter,gao2018black,liang2017deep,zhao2017generating}.
Overall \deleted{we have}\added{there are} two aims when generating the adversarial examples: (1) the examples should be incorrectly classified; and (2) the examples should be as similar to the original text as possible.

\begin{figure}[tb]
    \centering
    \includegraphics[width=\columnwidth]{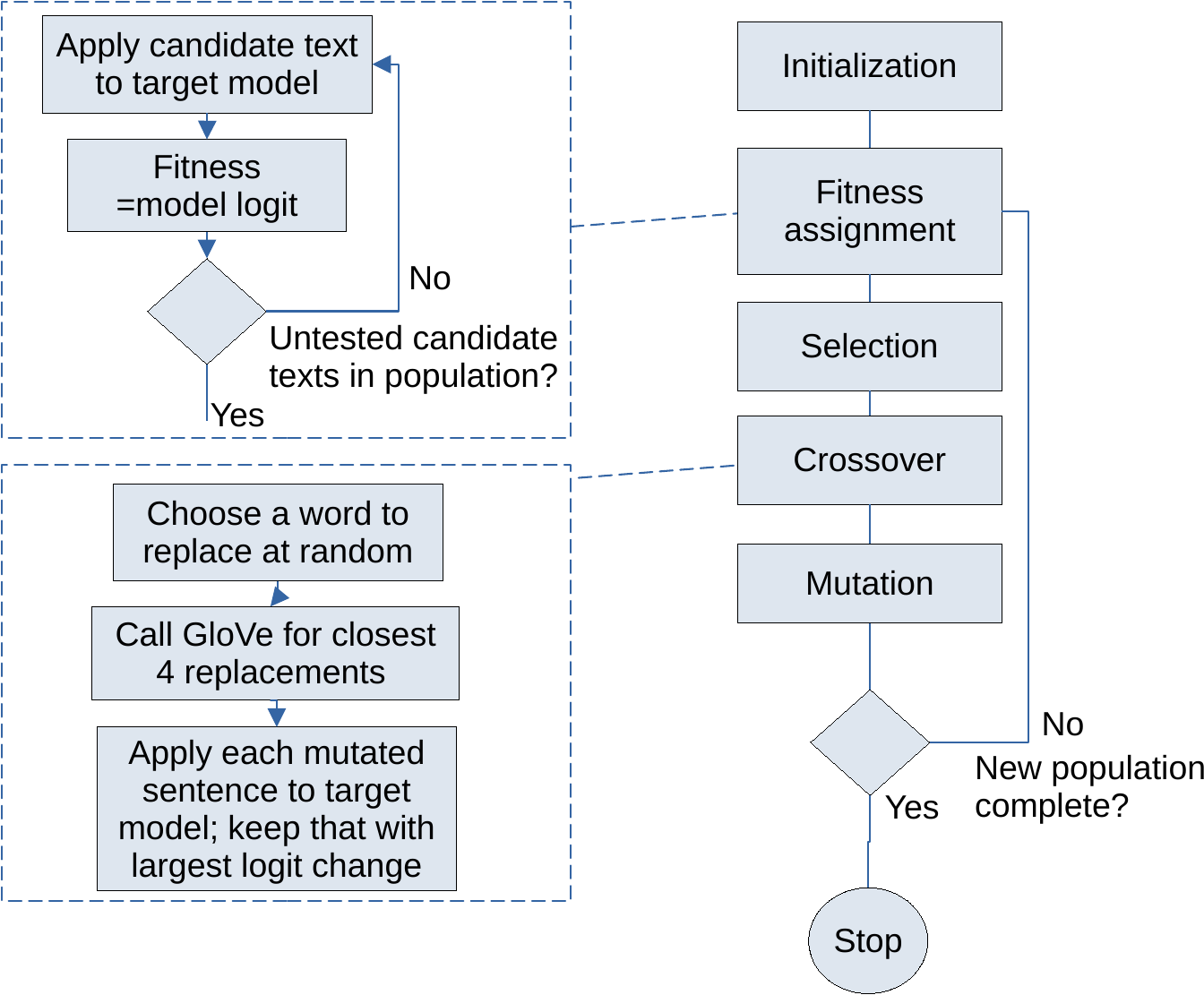}
    \caption{GA Workflow}
    \label{fig:ga-workflow}
\end{figure}

\deleted{We propose}\added{The proposal is} to use a GA to evolve a population of adversarial examples. GAs are a search-based framework for optimisation and have several components that can be constrained to guide the search to meet the aims noted above. GAs find a solution(s) based on the concept of \textit{fitness}, i.e., a relative measure of quality comparing one solution to another. GAs maintain a population of solutions for the problem at hand, and in each iteration \textit{select} fitter solutions, which are \textit{recombined} (a.k.a. crossover) and \textit{mutated} to generate a new population of solutions. This new population replaces the original one and the process repeats until some termination criterion is met: usually a solution of sufficient quality is found or a maximum number of solutions have been evaluated. 
The components listed above are brought together into an algorithm as illustrated in the workflow in \Cref{fig:ga-workflow}. \added{This figure is annotated with the sub-steps for the two stages that interact with the model, described shortly.} 
\deleted{In the}The remainder of this section \deleted{we} will describe \deleted{our}\added{the} implementation of the major components of the GA for generating adversarial examples. \deleted{We refer to the}\added{The} overall framework \added{is referred to} as \algname.

\deleted{We will aim}\added{The aim is} to maintain the grammatical and syntactical structure of the input text, \deleted{and so will be}\added{so} attempting to not only replace words with their synonyms, but to minimise the number of words replaced. The concept of fitness and the implementation of each operator is designed to reflect this strategy. \deleted{We take a}\added{A} black-box approach \deleted{and assume we have} \added{is taken that assumes} access to only the input and output of the model. \deleted{Our approach}\added{\algname} was implemented within the TextAttack framework, and the code artefact is publicly available \cite{artefact}.

\subsection{Representation}
\begin{figure}
    \centering
    \includegraphics[width=0.4\textwidth]{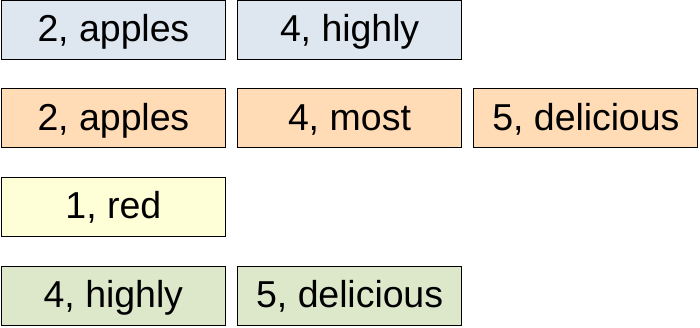}
    \caption{Four solutions using the compact representation. Each solution is a different colour for clarity. Solutions are a list of replacements for words in the original text: in this example the text was ``green grapes are very tasty'' so, for example, the top solution represents the adversarial text ``green apples are highly tasty''.}
    \label{fig:representation}
\end{figure}

An adversarial example is simply a variation of a piece of original text. The GA's job is to find such examples: so each ``solution'' in the population is one possible variation of the text. As such \deleted{we could choose to represent} each example \added{could be represented} as the full text but this is obviously rather wasteful of memory and, as crossover and mutation operations are applied to generate new solutions, could be computationally costly as well.
Consequently \deleted{we adopt} a compact representation for solutions \added{is adopted} that only stores the \textit{changes} to be made to the original text to make the adversarial example.
Each solution is a list of (location, replacement) pairs specifying which word in the original text shall be replaced and what word to replace it with. Solutions may be any length from zero (unchanged) up to the number of words in the original text (completely replacing the phrase). Some examples are given in \Cref{fig:representation}. This representation is less computationally intensive, and easier to manipulate, than storing the full mutated variants of the text.

\subsection{Initialization}
The GA begins by generating an initial population $P_0$ of size $S$. To generate each member of $P_0$, \deleted{we start}\added{the algorithm starts} with an empty solution (i.e., no change to the original text) and applies the mutation operator described in \Cref{sec:3mutations} five times.
Consequently, each solution in $P_0$ represents up to five word changes to the original text.
Following some preliminary tuning experiments, \deleted{we set}\added{the experiments had}  $S=60$\deleted{ in our experiments}.
The solutions in $P_0$ will then be evaluated by applying the word replacements to the original text, and passing the text to the NLP model for classification. The fitness for each solution is then determined as described in \Cref{sec:fitness}.

\subsection{Fitness and Selection}
\label{sec:fitness}
The fitness measure targets the first aim: incorrect classification, and is follows the default TextAttack Untargeted Classification implementation.
\deleted{We have}\added{There are} only two possible classifications for \deleted{our}\added{the} text: negative, and positive. The NLP model assigns a score to each class, with the class having the higher score being the prediction. The proximity to misclassification is simply the difference between these two scores, if the true class still has the higher score. The algorithm seeks to minimise this difference until the correct class no longer has the higher score.

A fitness proportionate selection \citep{luke2009essentials} operation uses this concept of fitness to choose parents for the remaining GA operators. Each solution is allocated a probability of selection, by taking its fitness divided by the total fitness of all solutions in the population. Solutions are then selected according to this probability so that they are chosen roughly in proportion with their fitness. \deleted{We use }TextAttack's implementation of fitness proportionate selection \added{is used}, which uses a softmax function to normalise the probability when sampling parents.

\subsection{Recombination / Crossover}
Two parent solutions chosen by selection are recombined into a single new offspring following a simple uniform crossover \citep{luke2009essentials}. Each replacement from the two parents has a 50\% probability of being copied into the offspring; where the two parents both have a replacement for one location in the text, one is chosen, with an equal probability of the either parent's replacement being copied.
Recombination is repeated until $S$ solutions are created (i.e., enough to replace the present population).

\subsection{Mutation}
\label{sec:3mutations}

Each solution within the GA's population is a set of individual word replacements.
\deleted{Our}\added{The} mutation operator is designed to allow for exploration of new parts of the search space, while achieving \deleted{our}\added{the} second overall aim by guiding the search towards examples semantically similar to the original text.

The operator simply adds a word replacement to the set of replacements in the solution.
A target word $w$ for replacement is selected uniformly at random from the original text, excluding the default NLTK stop-words (e.g., the, and, to, a ...).
If $w$ is already marked for replacement in the solution, the existing replacement is updated.

\begin{figure*}
    \centering
    \includegraphics[width=\textwidth]{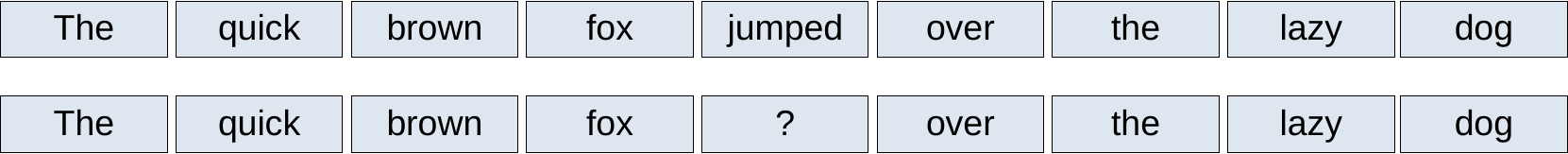}
    \caption{Masking is used in \deleted{our}\added{the} mutation operator, so the GloVe can identify the best fitting replacement words for a given context. In this case, the GloVe embedding is estimated for the location `?' and the words with the closest GloVe embeddings in Euclidean space to that target are selected as possible replacements for `jumped'.}
    \label{fig:mask}
\end{figure*}

To choose a replacement for $w$, a set of ``most likely'' replacements for $w$ is identified. Similar to \cite{ye2021heuristic,alzantot2018generating} \deleted{we used}\added{\algname uses} the pre-trained counter-fitted Global Vectors for Word Representation (GloVe) embedding \cite{mrkvsic2016counter} that is included with the TextAttack library\footnote{Specifically the \lstinline|WordEmbedding.counterfitted_GLOVE_embedding()| function} to translate the words into numeric vectors. 
GloVe \cite{pennington2014glove} tries to capture the sense that certain words are more likely to be seen in context of others. For example, ``lettuce'' is likely to be seen in the context of ``salad'' but not with ``train''. Context is implied through measuring the spatial distance between these word vectors, whereby the shorter the distance between word vectors the stronger the relationship between the actual words.
In the approach of \cite{ye2021heuristic,alzantot2018generating}, $w$ was converted to a vector using GloVe, then the closest words to $w$ in the Euclidean space served as possible replacements. \deleted{In our approach, we apply}\added{\algname applies} a mask. The four words (or fewer if near an end of the text) before and after $w$ are provided as the local context window for GloVe (\Cref{fig:mask}); $w$ is masked and the closest \textit{max\_neighbours} to the GloVe embedding for the space left by $w$ are determined. 
The final choice of word to replace $w$ is the one that takes the model prediction furthest toward the target label, given all the other replacements already applied.
The idea is that this will lead to more plausible word replacements that fit better with the surrounding context.


\subsection{Stopping Criteria}
\deleted{We have}\added{There are} two possible exit conditions for the algorithm. The first is the discovery of a population member that, after querying the sentiment model, is equal to the label and so \deleted{we have}\added{is} an adversarial example. 
The second condition is that \deleted{we have}\added{the algorithm has} run for a set number of generations. 
The maximum number of generations was determined after experimentation.
\deleted{We initially set}\added{Initially,} the number of generations \added{was set} at 100. After multiple trials \deleted{we}\added{it was} noted that after 20 generations there was no improvement in the adversarial examples generated. 

\section{Experiments and Results} 
\label{sec:exp}
\label{sec:experiments}
\subsection{Evaluation Metrics}

\deleted{In our results, we report several measures}\added{Several measures are used} to compare the approaches. When the model is applied to the test set, \deleted{we have} the standard measures TP, TN, FP, FN \added{are used} for true/false positive/negative classifications, with a bar (e.g., $\overline{TP}$) for these measures when applied to the attacked test set. The measures are defined as:
\begin{description}
    \item[Accuracy] Standard classification accuracy; \% of correct predictions on the test set after attack has been applied; 
    \begin{equation}
        Acc = \frac{100(\overline{TP} + \overline{TN})}{\overline{TP} + \overline{TN} + \overline{FP} + \overline{TN}}
    \end{equation}
    \item[Original Accuracy] Standard classification accuracy; \% of correct predictions on the original unaltered test set; 
    \begin{equation}
        Acc_{org} = \frac{100(TP + TN)}{TP + TN + FP + TN}
    \end{equation}
    \item[\% Perturbed Words] Mean (over test set) \% of words changed from original in each adversarial example; if the test set $S$ contains $|S|$ documents, each document $s_i \in S$ in the original test set contains $|s_i|$ words, and $H(s_i,\overline{s_i})$ is the Hamming distance between the original document $s_i$ and the corresponding adversarial example $\overline{s_i}$, then this measure is: 
    \begin{equation}
        PW = \frac{\sum_{i=1}^{|S|}\frac{100H(s_i,\overline{s_i})}{s_i}}{|S|}
    \end{equation}
    \item[Semantic Similarity] Mean (over test set) USE similarity between original text and adversarial example: 
    \begin{equation}
        SS = \sum_{i=1}^{|S|}USE(s_i,\overline{s_i})
    \end{equation}
    \item[Query Number] $QN$ The number of queries of target model to generate adversarial example set
    \item[Computation Time] $CT$ Wall-clock time to generate adversarial example set
\end{description}

\subsection{Experimental Settings}

Using \textit{TextAttack} framework \cite{morris2020textattack}, \deleted{We compare} the proposed algorithm \added{is compared} against the state-of-art methods (BAE \cite{garg2020bae}, A2T \cite{yoo2021towards}), which are the best candidates to compare against in terms of black-box text replacement. The following hardware specifications are used: GPU (\verb|GeForce RTX 2080|), CPU (\verb|Intel(R) Core(TM) i9-9900K|), and RAM (\verb|16GB|). To validate the performance of the proposed work, two popular datasets were used: (1) Movie Reviews (MR) \cite{Pang2005SeeingSE}: binary classification on movie reviews (positive, negative) with data split of 9K train \& 1K test samples and its average sentence length is 20 words, (2) AG-News  \cite{zhang2015character}: multi-class classification on news articles (business, science, sports, world) with data split of 120K train and 7.6K test samples and its average sentence length is 43 words. Besides that, three common text classification models were used:  WordCNN \cite{Kim2014ConvolutionalNN}, WordLSTM \cite{hochreiter1997long}, BERT \cite{devlin2018bert}. These pretrained models are all available in the TextAttack framework. The major configuration parameters of the genetic algorithm are detailed in \Cref{tab:configuration}; the full implementation of \algname is available in the paper's accompanying artefact \cite{artefact}.



\begin{table}[tb]
    \centering
    \begin{tabular}{l|c}
        Parameter & Value \\ \hline
         Max generations & 30 \\
                             Population size & 20 \\
                             Perturbations & 5 \\
                             Temperature for softmax used selection & 0.3 \\
                             Crossover Rate & 100\% \\
    \end{tabular}
    \caption{Major configuration settings for \algname}
    \label{tab:configuration}
\end{table}

\subsection{Quantitative results}
\Cref{tab:perf-mr,tab:perf-agn} report the performance of the three models on the original test data and the test data with each attack applied. All the approaches reduce classification accuracy by a substantial amount on both data sets and with all three models. BAE performs slightly better than A2T, and \algname reduces accuracy further still.

However, this reduction in accuracy comes at a cost. \algname typically perturbs just under twice as many words as the other two methods. Semantic similarity of the text compared to the original is also slightly lower with \algname than with the other methods, and \algname requires more queries of the model (just under 2x that of BAE and 6--8x that of A2T). Thus \algname adds to the trade-off: for slightly more change to the text, the explorative power of the GA is able to greatly reduce accuracy.

The additional model queries do not feed through to greatly increased run times for \algname. \Cref{tab:eval-time-mr} shows the wall clock run times in seconds for the three approaches. \algname is the fastest of the three approaches on both the WordCNN and WordLSTM models, and approximately 5\% longer running then BAE on BERT.

\added{In practical terms, \algname is preferable where effectiveness is critical, when trying to maximally expose model vulnerability and where text fidelity is less important. This could include security evaluation contexts for high-stakes domains (e.g., finance, law, medicine) where maximum robustness testing may be preferred even if perturbations are unrealistic. In such offline evaluation settings the additional number of queries does not matter as much; but this could in future be tackled by adding early stopping or response caching.
}

\begin{table}[]
\centering
\begin{tabular}{@{}llcccc@{}}
\toprule
\multirow{2}{*}{Models} & \multirow{2}{*}{Metric} & \multirow{2}{*}{\begin{tabular}[c]{@{}c@{}}$Acc_{org}$\end{tabular}} & \multicolumn{3}{c}{Adv Attack} \\ \cmidrule(l){4-6} 
 &  &  & BAE & A2T & \algname \\ \midrule
\multirow{4}{*}{WordCNN} & $Acc$ & \multirow{4}{*}{76.83} & 27.58 & 44.28 & 5.82 \\
 & $PW$ &  & 12.85 & 13.68 & 21.37 \\
 & $SS$ &  & 0.85 & 0.86 & 0.83 \\
 & $QN$ &  & 59.25 & 15.56 & 91.72 \\ \midrule
\multirow{4}{*}{WordLSTM} & $Acc$ & \multirow{4}{*}{77.86} & 24.95 & 38.56 & 7.04 \\
 & $PW$ &  & 11.75 & 11.37 & 20.42 \\
 & $SS$ &  & 0.86 & 0.88 & 0.84 \\
 & $QN$ &  & 57.23 & 14.45 & 88.83 \\ \midrule
\multirow{4}{*}{BERT} & $Acc$ & \multirow{4}{*}{84.24} & 36.49 & 59.76 & 19.51 \\
 & $PW$ &  & 13.73 & 12.03 & 20.74 \\
 & $SS$ &  & 0.85 & 0.87 & 0.84 \\
 & $QN$ &  & 63.07 & 17.73 & 132.63 \\ \bottomrule
\end{tabular}
\caption{Performance evaluation - MR}
\label{tab:perf-mr}
\end{table}

\begin{table}[]
\centering
\begin{tabular}{@{}llcccc@{}}
\toprule
\multirow{2}{*}{Models} & \multirow{2}{*}{Metric} & \multirow{2}{*}{\begin{tabular}[c]{@{}c@{}}$Acc_{org}$\end{tabular}} & \multicolumn{3}{c}{Adv Attack} \\ \cmidrule(l){4-6} 
 &  &  & BAE & A2T & \algname \\ \midrule
\multirow{4}{*}{WordCNN} & $Acc$ & \multirow{4}{*}{91.57} & 73.93 & 82.22 & 56.89 \\
 & $PW$ &  & 6.66 & 8.52 & 13.10 \\
 & $SS$ &  & 0.92 & 0.91 & 0.88 \\
 & $QN$ &  & 113.6 & 26.96 & 203.79 \\ \midrule
\multirow{4}{*}{WordLSTM} & $Acc$ & \multirow{4}{*}{91.63} & 73.53 & 77.49 & 62.08 \\
 & $PW$ &  & 6.69 & 7.59 & 12.74 \\
 & $SS$ &  & 0.92 & 0.91 & 0.88 \\
 & $QN$ &  & 115.96 & 24.6 & 204.91 \\ \midrule
\multirow{4}{*}{BERT} & $Acc$ & \multirow{4}{*}{95.14} & 81.55 & 82.86 & 73.17 \\
 & $PW$ &  & 7.28 & 7.70 & 12.26 \\
 & $SS$&  & 0.93 & 0.91 & 0.88 \\
 & $QN$ &  & 122.63 & 27.23 & 218.45 \\ \bottomrule
\end{tabular}
\caption{Performance evaluation - AG-News}
\label{tab:perf-agn}
\end{table}

\begin{table}[]
\centering
\begin{tabular}{@{}cccc@{}}
\toprule
\multirow{2}{*}{Adv Attack} & \multicolumn{3}{c}{Models} \\ \cmidrule(l){2-4} 
 & WordCNN & WordLSTM & BERT \\ \midrule
BAE & 666 & 492 & 1484 \\
A2T & 117 & 116 & 400 \\
\algname & 66 & 98 & 1547 \\ \bottomrule
\end{tabular}
\caption{Computation time (in seconds) over test sub-set of MR}
\label{tab:eval-time-mr}
\end{table}

\subsection{Qualitative results}
\deleted{We include }Illustrative examples of the output generated by each method \added{are given} in \Cref{tab:eval-text-MR-BERT,tab:eval-text-MR-CNN-1,tab:eval-text-MR-CNN-2,tab:eval-text-AGN-BERT,tab:eval-text-AGN-CNN}. The extent of changes made by \algname is often more than that of the other methods, but typically more subtle, with fewer changes to the literal meaning of the original. 

With the example for the BERT model on the MR data (\Cref{tab:eval-text-MR-BERT}), \algname used two additional replacements than BAE; but BAE changed ``portray'' to the less natural ``appear'' rather than ``describes''; BAE also replaced ``powerful'' with ``short'', rather than the much closer ``emphatic''. Similarly \algname did not replace ``filmmaking'' with ``cinematographic'' as A2T did. Nevertheless, in this example \algname's replacements were less grammatically sound than those of BAE; though remained comparable with A2T. On the WordCNN classifier with the same data (\Cref{tab:eval-text-MR-CNN-1,tab:eval-text-MR-CNN-2}), \algname replaced ``tears'' with the more appropriate ``crying'' rather than ``greed'', replaced ``melodrama'' with ``melodramadic'' vs ``shame'', and replaced ``mr'' with ``mister'' rather than ``love''. However, \algname also replaced ``the'' with ``per'': a mistake that was observed several times in the full results for both \algname and A2T. In general the examples featuring comma separated lists rather than full sentences seemed to lead to better results for all attacks, perhaps because the context matters less in these cases.

For BERT on the AG-News set (\Cref{tab:eval-text-AGN-BERT}), \algname uses one more substitution than BAE: but both are subtle: replacement of ``Congress'' with ``Senate'' rather than an insertion to create ``Congress senate'', and replacement of ``final'' with ``ultimate'', which works well in this context. \algname uses fewer substitutions than A2T, which also made a factual change by replacing ``Wednesday'' with ``Thursday''. For WordCNN on AG-News (\Cref{tab:eval-text-AGN-CNN}), \algname replaced ``Games'' with ``Gaming'' rather than the less appropriate ``time'' (common to both BAE and A2T). \algname did introduce an additional change: ``unclear'' was replaced with the similar ``nebulous''.

In general, the adversarial examples from all methods replaced proper nouns and abbrevations, e.g. ag-news ``T N'' being replaced with random other initials and ``Melbourne'' being replaced with other Australian cities. This is a possible flaw in only using similarity in the embedding space. Clearly proper nouns such as these are used in similar ways across a corpus so will have similar embeddings, but are not interchangeable. \added{Named Entity Recognition \cite{nasar2021named} represents a suitable direction for future research to address this issue.}



\begin{table}[]
\centering
\begin{tabular}{@{}cl@{}}
\toprule
Attack & \multicolumn{1}{c}{Text} \\ \midrule
Original & \begin{tabular}[c]{@{}l@{}}it uses some of the \textcolor{blue}{figures} from the real-life story to \\ \textcolor{blue}{portray} themselves in the film . the result is a \textcolor{blue}{powerful} , \\ \textcolor{blue}{naturally dramatic} piece of low-budget \textcolor{blue}{filmmaking} .\end{tabular} \\
\midrule
BAE & \begin{tabular}[c]{@{}l@{}}it uses some of the \textcolor{cyan}{figures} from the real-life story to \\ \textcolor{red}{appear} themselves in the film . the result is a \textcolor{red}{short} , \\ \textcolor{cyan}{naturally dramatic} piece of low-budget \textcolor{cyan}{filmmaking} .\end{tabular} \\
\midrule
A2T & \begin{tabular}[c]{@{}l@{}}it uses some of the \textcolor{cyan}{figures} from the real-life story to \\ \textcolor{cyan}{portray} themselves in the film . the result is a \textcolor{red}{emphatic} ,  \\ \textcolor{red}{evidently} \textcolor{red}{prodigious} piece of low-budget \textcolor{red}{cinematographic} .\end{tabular} \\
\midrule
\algname & \begin{tabular}[c]{@{}l@{}}it uses some of the \textcolor{red}{numerals} from the real-life story to \\ \textcolor{red}{describes} themselves in the film . the result is a \textcolor{red}{emphatic} ,  \\ \textcolor{cyan}{naturally} \textcolor{red}{whopping} piece of low-budget \textcolor{cyan}{filmmaking} .\end{tabular} \\ \bottomrule
\end{tabular}
\caption{Qualitative examples of \algname attacks against BAE and A2T on the BERT classifier over MR dataset. ({\color{blue} Blue}: modification candidate, {\color{red} Red} and {\color{cyan} Cyan}: modified and non-modified words for each attack.)}
\label{tab:eval-text-MR-BERT}
\end{table}


\begin{table}[]
\centering
\begin{tabular}{@{}cl@{}}
\toprule
Attack & \multicolumn{1}{c}{Text} \\ \midrule
Original & \begin{tabular}[c]{@{}l@{}}
\textcolor{blue}{passion} , \textcolor{blue}{melodrama} , \textcolor{blue}{sorrow} , laugther , and \textcolor{blue}{tears} cascade \\ over \textcolor{blue}{the} \textcolor{blue}{screen} effortlessly . . . 
\end{tabular} \\
\midrule
BAE & \begin{tabular}[c]{@{}l@{}}
\textcolor{red}{passion} , \textcolor{red}{shame} , \textcolor{red}{humor} , laugther , and \textcolor{red}{greed} cascade \\ over \textcolor{cyan}{the} \textcolor{red}{pages} effortlessly . . . 
\end{tabular} \\
\midrule
A2T & \begin{tabular}[c]{@{}l@{}}
\textcolor{red}{fervor} , \textcolor{cyan}{melodrama} , \textcolor{red}{bereavement} , laugther , and \textcolor{red}{sobs} cascade \\ over \textcolor{cyan}{the} \textcolor{cyan}{screen} effortlessly . . . 
\end{tabular} \\
\midrule
\algname & \begin{tabular}[c]{@{}l@{}}
\textcolor{cyan}{passion} , \textcolor{red}{melodramatic} , \textcolor{cyan}{sorrow} , laugther , and \textcolor{red}{crying} cascade \\ over \textcolor{red}{per} \textcolor{cyan}{screen} effortlessly . . . 
\end{tabular} \\ \bottomrule
\end{tabular}
\caption{Qualitative examples of \algname attacks against BAE and A2T on the WordCNN classifier over MR dataset. ({\color{blue} Blue}: modification candidate, {\color{red} Red} and {\color{cyan} Cyan}: modified and non-modified words for each attack.)}
\label{tab:eval-text-MR-CNN-1}
\end{table}


\begin{table}[]
\centering
\begin{tabular}{@{}cl@{}}
\toprule
Attack & \multicolumn{1}{c}{Text} \\ \midrule
Original & \begin{tabular}[c]{@{}l@{}}
\textcolor{blue}{skillful} as he is , \textcolor{blue}{mr} . shyamalan is \textcolor{blue}{undone} \\ by his pretensions . 
\end{tabular} \\
\midrule
BAE & \begin{tabular}[c]{@{}l@{}}
\textcolor{cyan}{skillful} as he is , \textcolor{red}{love} . shyamalan is \textcolor{cyan}{undone} \\ by his pretensions . 
\end{tabular} \\
\midrule
A2T & \begin{tabular}[c]{@{}l@{}}
\textcolor{red}{crafty} as he is , \textcolor{red}{mister} . shyamalan is \textcolor{cyan}{undone} \\ by his pretensions . 
\end{tabular} \\
\midrule
\algname & \begin{tabular}[c]{@{}l@{}}
\textcolor{cyan}{skillful} as he is , \textcolor{red}{mister} . shyamalan is \textcolor{red}{conquered} \\ by his pretensions .
\end{tabular} \\ \bottomrule
\end{tabular}
\caption{Qualitative examples of \algname attacks against BAE and A2T on the WordCNN classifier over MR dataset. ({\color{blue} Blue}: modification candidate, {\color{red} Red} and {\color{cyan} Cyan}: modified and non-modified words for each attack.)}
\label{tab:eval-text-MR-CNN-2}
\end{table}


\begin{table}[]
\centering
\begin{tabular}{@{}cl@{}}
\toprule
Attack & \multicolumn{1}{c}{Text} \\ \midrule
Original & \begin{tabular}[c]{@{}l@{}}
\textcolor{blue}{Congress Passes Bill}  Allowing Space \textcolor{blue}{Tours} (AP) AP\\ -\textcolor{blue}{Outer} space could become the \textcolor{blue}{final} frontier of \\ tourism under legislation passed \textcolor{blue}{Wednesday} by the \\ Senate to regulate commercial human spaceflight.
\end{tabular} \\
\midrule
BAE & \begin{tabular}[c]{@{}l@{}}
\textcolor{cyan}{Congress} \textcolor{red}{senate} \textcolor{cyan}{Bill} Allowing Space \textcolor{cyan}{Tours}  (AP) AP \\-\textcolor{cyan}{Outer} space could become the \textcolor{cyan}{final} frontier  of \\tourism under legislation passed \textcolor{cyan}{Wednesday} by the \\ Senate to regulate commercial human spaceflight.
\end{tabular} \\
\midrule
A2T & \begin{tabular}[c]{@{}l@{}}
\textcolor{cyan}{Congress} \textcolor{cyan}{Passes} \textcolor{red}{Bills} Allowing Space \textcolor{red}{Visits}  (AP) AP \\-\textcolor{red}{Outside} space could become the \textcolor{cyan}{final} frontier  of \\ tourism under legislation passed \textcolor{red}{Thursday} by the \\ Senate to regulate commercial human spaceflight.
\end{tabular} \\
\midrule
\algname & \begin{tabular}[c]{@{}l@{}}
\textcolor{red}{Senate} \textcolor{cyan}{Passes} \textcolor{cyan}{Bill} Allowing Space \textcolor{cyan}{Tours}  (AP) AP \\ -\textcolor{cyan}{Outer} space could become the \textcolor{red}{ultimate} frontier  of \\ tourism under legislation passed \textcolor{cyan}{Wednesday} by the \\ Senate to regulate commercial human spaceflight.
\end{tabular} \\ \bottomrule
\end{tabular}
\caption{Qualitative examples of \algname attacks against BAE and A2T on the BERT classifier over AG-News dataset. ({\color{blue} Blue}: modification candidate, {\color{red} Red} and {\color{cyan} Cyan}: modified and non-modified words for each attack.)}
\label{tab:eval-text-AGN-BERT}
\end{table}


\begin{table}[]
\centering
\begin{tabular}{@{}cl@{}}
\toprule
Attack & \multicolumn{1}{c}{Text} \\ \midrule
Original & \begin{tabular}[c]{@{}l@{}}
A \#39;new Greece \#39; beams after success of \textcolor{blue}{Games} As \\ Greeks get a boost, it remains \textcolor{blue}{unclear} if \textcolor{black}{success} will \textcolor{black}{mean} \\ higher stature in Europe. By Peter Ford \textcolor{black}{Staff} \textcolor{black}{writer} of The \\ Christian Science Monitor. 
\end{tabular} \\
\midrule
BAE & \begin{tabular}[c]{@{}l@{}}
A \#39;new Greece \#39; beams after success of \textcolor{red}{time} As \\ Greeks get a boost, it remains \textcolor{cyan}{unclear} if \textcolor{black}{success} will \textcolor{black}{mean} \\ higher stature in Europe. By Peter Ford \textcolor{black}{Staff} \textcolor{black}{writer} of The \\ Christian Science Monitor. 
\end{tabular} \\
\midrule
A2T & \begin{tabular}[c]{@{}l@{}}
A \#39;new Greece \#39; beams after success of \textcolor{red}{time} As \\ Greeks get a boost, it remains \textcolor{cyan}{unclear} if \textcolor{black}{success} will \textcolor{black}{mean} \\ higher stature in Europe. By Peter Ford \textcolor{black}{Staff} \textcolor{black}{writer} of The \\ Christian Science Monitor. 
\end{tabular} \\
\midrule
\algname & \begin{tabular}[c]{@{}l@{}}
A \#39;new Greece \#39; beams after success of \textcolor{red}{Gaming} As \\ Greeks get a boost, it remains \textcolor{red}{nebulous} if \textcolor{black}{success} will \textcolor{black}{mean} \\ higher stature in Europe. By Peter Ford \textcolor{black}{Staff} \textcolor{black}{writer} of The \\ Christian Science Monitor. 
\end{tabular} \\ \bottomrule
\end{tabular}
\caption{Qualitative examples of \algname attacks against BAE and A2T on the WordCNN classifier over AG-News dataset. ({\color{blue} Blue}: modification candidate, {\color{red} Red} and {\color{cyan} Cyan}: modified and non-modified words for each attack.)}
\label{tab:eval-text-AGN-CNN}
\end{table}

\section{Conclusion}
\label{sec:conclusion}
\deleted{We have}\added{This paper has} proposed \algname, a hybrid approach to generating adversarial text seeking the best of previous work: a search based approach using a genetic algorithm that only needs the logit value output from the model, guided by GloVe for semantic similarity. In practice \algname reduces accuracy even more than the BAE and A2T approaches \deleted{we compared against}, but at the cost of needing 2x-8x as many queries of the target model and perturbing around twice as many words on average. However, the resulting examples show only a slight reduction in semantic similarity over those generated by the other approaches, and only around 5\% increase in run time despite the greater number of model queries.

There are several lines of potential future work. \deleted{We}\added{The algorithm} could draw further on BERT to use the context of words being perturbed to further improve the semantic similarity of the replacement text. \deleted{We might also apply }A simple local search \added{could be applied} during each replacement, taking the top-n most likely words according to BERT given the word's context, testing the resulting text for each replacement before deciding which one to keep (this is known as a memetic algorithm \cite{moscato1992memetic}). Amending the objective function for the GA to include a penalty for replacements could also improve the similarity of the adversarial examples to the original text.
It would be interesting to try to identify where proper nouns and abbreviations appear to prevent their replacement. 
Clearly there is also scope for drawing from more recent large language models i.e. LLaMA \cite{grattafiori2024llama}, Mixtral \cite{jiang2024mixtral}), but at a potential increase in computational cost.






\bibliographystyle{unsrt} 
\bibliography{refs}

@article{jiang2024mixtral,
  title={Mixtral of experts},
  author={Jiang, Albert Q and Sablayrolles, Alexandre and Roux, Antoine and Mensch, Arthur and Savary, Blanche and Bamford, Chris and Chaplot, Devendra Singh and Casas, Diego de las and Hanna, Emma Bou and Bressand, Florian and others},
  journal={arXiv preprint arXiv:2401.04088},
  year={2024}
}

@article{grattafiori2024llama,
  title={The llama 3 herd of models},
  author={Grattafiori, Aaron and Dubey, Abhimanyu and Jauhri, Abhinav and Pandey, Abhinav and Kadian, Abhishek and Al-Dahle, Ahmad and Letman, Aiesha and Mathur, Akhil and Schelten, Alan and Vaughan, Alex and others},
  journal={arXiv preprint arXiv:2407.21783},
  year={2024}
}

@article{sanh2019distilbert,
  title={DistilBERT, a distilled version of BERT: smaller, faster, cheaper and lighter},
  author={Sanh, Victor and Debut, Lysandre and Chaumond, Julien and Wolf, Thomas},
  journal={arXiv preprint arXiv:1910.01108},
  year={2019}
}

@article{liu2019roberta,
  title={Roberta: A robustly optimized bert pretraining approach},
  author={Liu, Yinhan and Ott, Myle and Goyal, Naman and Du, Jingfei and Joshi, Mandar and Chen, Danqi and Levy, Omer and Lewis, Mike and Zettlemoyer, Luke and Stoyanov, Veselin},
  journal={arXiv preprint arXiv:1907.11692},
  year={2019}
}

@inproceedings{devlin2019bert,
  title={Bert: Pre-training of deep bidirectional transformers for language understanding},
  author={Devlin, Jacob and Chang, Ming-Wei and Lee, Kenton and Toutanova, Kristina},
  booktitle={Proceedings of the 2019 conference of the North American chapter of the association for computational linguistics: human language technologies, volume 1 (long and short papers)},
  pages={4171--4186},
  year={2019}
}

@article{gowal2018effectiveness,
  title={On the effectiveness of interval bound propagation for training verifiably robust models},
  author={Gowal, Sven and Dvijotham, Krishnamurthy and Stanforth, Robert and Bunel, Rudy and Qin, Chongli and Uesato, Jonathan and Arandjelovic, Relja and Mann, Timothy and Kohli, Pushmeet},
  journal={arXiv preprint arXiv:1810.12715},
  year={2018}
}

@article{Szegedy2013IntriguingPO,
  title={Intriguing properties of neural networks},
  author={Christian Szegedy and Wojciech Zaremba and Ilya Sutskever and Joan Bruna and D. Erhan and Ian J. Goodfellow and Rob Fergus},
  journal={CoRR},
  year={2013},
  volume={abs/1312.6199},
  url={https://api.semanticscholar.org/CorpusID:604334}
}

@inproceedings{jia2019certified,
  title={Certified Robustness to Adversarial Word Substitutions},
  author={Jia, Robin and Raghunathan, Aditi and G{\"o}ksel, Kerem and Liang, Percy},
  booktitle={Proceedings of the 2019 Conference on Empirical Methods in Natural Language Processing and the 9th International Joint Conference on Natural Language Processing (EMNLP-IJCNLP)},
  pages={4129--4142},
  year={2019}
}

@inproceedings{jin2020bert,
  title={Is bert really robust? a strong baseline for natural language attack on text classification and entailment},
  author={Jin, Di and Jin, Zhijing and Zhou, Joey Tianyi and Szolovits, Peter},
  booktitle={Proceedings of the AAAI conference on artificial intelligence},
  volume={34},
  number={05},
  pages={8018--8025},
  year={2020}
}

@inproceedings{wang2022semattack,
  title={SemAttack: Natural Textual Attacks via Different Semantic Spaces},
  author={Wang, Boxin and Xu, Chejian and Liu, Xiangyu and Cheng, Yu and Li, Bo},
  booktitle={Findings of the Association for Computational Linguistics: NAACL 2022},
  pages={176--205},
  year={2022}
}

@inproceedings{yuan2023bridge,
  title={Bridge the Gap Between CV and NLP! A Gradient-based Textual Adversarial Attack Framework},
  author={Yuan, Lifan and Zhang, Yichi and Chen, Yangyi and Wei, Wei},
  booktitle={Findings of the Association for Computational Linguistics: ACL 2023},
  pages={7132--7146},
  year={2023}
}

@inproceedings{li2023adversarial,
  title={Adversarial Text Generation by Search and Learning},
  author={Li, Guoyi and Shi, Bingkang and Liu, Zongzhen and Kong, Dehan and Wu, Yulei and Zhang, Xiaodan and Huang, Longtao and Lyu, Honglei},
  booktitle={The 2023 Conference on Empirical Methods in Natural Language Processing},
  year={2023}
}

@inproceedings{yu2022texthacker,
  title={TextHacker: Learning based Hybrid Local Search Algorithm for Text Hard-label Adversarial Attack},
  author={Yu, Zhen and Wang, Xiaosen and Che, Wanxiang and He, Kun},
  booktitle={Findings of the Association for Computational Linguistics: EMNLP 2022},
  pages={622--637},
  year={2022}
}

@inproceedings{ye2022texthoaxer,
  title={Texthoaxer: Budgeted hard-label adversarial attacks on text},
  author={Ye, Muchao and Miao, Chenglin and Wang, Ting and Ma, Fenglong},
  booktitle={Proceedings of the AAAI Conference on Artificial Intelligence},
  volume={36},
  number={4},
  pages={3877--3884},
  year={2022}
}

@inproceedings{ye2021heuristic,
  title={Heuristic-word-selection Genetic Algorithm for Generating Natural Language Adversarial Examples},
  author={Ye, Shijun and Zhang, Pengcheng and Dong, Hai and Ji, Shunhui},
  booktitle={2021 IEEE International Conference on Artificial Intelligence Testing (AITest)},
  pages={39--40},
  year={2021},
  organization={IEEE}
}

@article{angeles2020investor,
  title={Investor sentiment in the theoretical field of behavioural finance},
  author={{\'A}ngeles L{\'o}pez-Cabarcos, M and M P{\'e}rez-Pico, Ada and L{\'o}pez Perez, Maria Luisa and others},
  journal={Economic research-Ekonomska istra{\v{z}}ivanja},
  volume={33},
  number={1},
  pages={2101--2119},
  year={2020},
  publisher={Taylor and Francis Group i Sveu{\v{c}}ili{\v{s}}te Jurja Dobrile u Puli, Fakultet~…}
}

@article{lucey2005role,
  title={The role of feelings in investor decision-making},
  author={Lucey, Brian M and Dowling, Michael},
  journal={Journal of economic surveys},
  volume={19},
  number={2},
  pages={211--237},
  year={2005},
  publisher={Wiley Online Library}
}

@article{go2009twitter,
  title={Twitter sentiment classification using distant supervision},
  author={Go, Alec and Bhayani, Richa and Huang, Lei},
  journal={CS224N project report, Stanford},
  volume={1},
  number={12},
  pages={2009},
  year={2009}
}

@inproceedings{pennington2014glove,
  title={Glove: Global vectors for word representation},
  author={Pennington, Jeffrey and Socher, Richard and Manning, Christopher D},
  booktitle={Proceedings of the 2014 conference on empirical methods in natural language processing (EMNLP)},
  pages={1532--1543},
  year={2014}
}

@article{mrkvsic2016counter,
  title={Counter-fitting word vectors to linguistic constraints},
  author={Mrk{\v{s}}i{\'c}, Nikola and S{\'e}aghdha, Diarmuid O and Thomson, Blaise and Ga{\v{s}}i{\'c}, Milica and Rojas-Barahona, Lina and Su, Pei-Hao and Vandyke, David and Wen, Tsung-Hsien and Young, Steve},
  journal={arXiv preprint arXiv:1603.00892},
  year={2016}
}

@article{szegedy2013intriguing,
  title={Intriguing properties of neural networks},
  author={Szegedy, Christian and Zaremba, Wojciech and Sutskever, Ilya and Bruna, Joan and Erhan, Dumitru and Goodfellow, Ian and Fergus, Rob},
  journal={arXiv preprint arXiv:1312.6199},
  year={2013}
}

@inproceedings{gao2018black,
  title={Black-box generation of adversarial text sequences to evade deep learning classifiers},
  author={Gao, Ji and Lanchantin, Jack and Soffa, Mary Lou and Qi, Yanjun},
  booktitle={2018 IEEE Security and Privacy Workshops (SPW)},
  pages={50--56},
  year={2018},
  organization={IEEE}
}

@article{liang2017deep,
  title={Deep text classification can be fooled},
  author={Liang, Bin and Li, Hongcheng and Su, Miaoqiang and Bian, Pan and Li, Xirong and Shi, Wenchang},
  journal={arXiv preprint arXiv:1704.08006},
  year={2017}
}

@article{zhao2017generating,
  title={Generating natural adversarial examples},
  author={Zhao, Zhengli and Dua, Dheeru and Singh, Sameer},
  journal={arXiv preprint arXiv:1710.11342},
  year={2017}
}

@inproceedings{wang2019natural,
  title={Natural language adversarial attacks and defenses in word level},
  author={Wang, Xiaosen and Hao, Jin and He, Kun},
  booktitle={ArXiv preprint arXiv:1909.06723v1},
  year={2019},
  url={arXiv:1909.06723v1}
}

@article{hochreiter1997long,
  title={Long short-term memory},
  author={Hochreiter, Sepp and Schmidhuber, J{\"u}rgen},
  journal={Neural computation},
  volume={9},
  number={8},
  pages={1735--1780},
  year={1997},
  publisher={MIT Press}
}

@article{anderson1994genetic,
  title={Genetic algorithms for combinatorial optimization: the assemble line balancing problem},
  author={Anderson, Edward J and Ferris, Michael C},
  journal={ORSA Journal on Computing},
  volume={6},
  number={2},
  pages={161--173},
  year={1994},
  publisher={INFORMS}
}

@article{gebru2017using,
  title={Using deep learning and Google Street View to estimate the demographic makeup of neighborhoods across the United States},
  author={Gebru, Timnit and Krause, Jonathan and Wang, Yilun and Chen, Duyun and Deng, Jia and Aiden, Erez Lieberman and Fei-Fei, Li},
  journal={Proceedings of the National Academy of Sciences},
  volume={114},
  number={50},
  pages={13108--13113},
  year={2017},
  publisher={National Acad Sciences}
}

@inproceedings{dodge2017study,
  title={A study and comparison of human and deep learning recognition performance under visual distortions},
  author={Dodge, Samuel and Karam, Lina},
  booktitle={2017 26th international conference on computer communication and networks (ICCCN)},
  pages={1--7},
  year={2017},
  organization={IEEE}
}

@article{devlin2018bert,
  title={Bert: Pre-training of deep bidirectional transformers for language understanding},
  author={Devlin, Jacob and Chang, Ming-Wei and Lee, Kenton and Toutanova, Kristina},
  journal={arXiv preprint arXiv:1810.04805},
  year={2018}
}

@inproceedings{reuver-etal-2021-nlp,
    title = "No {NLP} Task Should be an Island: Multi-disciplinarity for Diversity in News Recommender Systems",
    author = "Reuver, Myrthe  and
      Fokkens, Antske  and
      Verberne, Suzan",
    booktitle = "Proceedings of the EACL Hackashop on News Media Content Analysis and Automated Report Generation",
    month = apr,
    year = "2021",
    address = "Online",
    publisher = "Association for Computational Linguistics",
    url = "https://aclanthology.org/2021.hackashop-1.7",
    pages = "45--55",
}

@inproceedings{alzantot2018generating,
  title={Generating Natural Language Adversarial Examples},
  author={Alzantot, Moustafa and Sharma, Yash and Elgohary, Ahmed and Ho, Bo-Jhang and Srivastava, Mani and Chang, Kai-Wei},
  booktitle={Proceedings of the 2018 Conference on Empirical Methods in Natural Language Processing},
  pages={2890--2896},
  year={2018}
}

@inproceedings{li2020bert,
  title={BERT-ATTACK: Adversarial Attack against BERT Using BERT},
  author={Li, Linyang and Ma, Ruotian and Guo, Qipeng and Xue, Xiangyang and Qiu, Xipeng},
  booktitle={Proceedings of the 2020 Conference on Empirical Methods in Natural Language Processing (EMNLP)},
  pages={6193--6202},
  year={2020}
}

@book{bennis2020nature,
  title={Nature-Inspired Methods for Metaheuristics Optimization: Algorithms and Applications in Science and Engineering},
  author={Bennis, Fouad and Bhattacharjya, Rajib Kumar},
  volume={16},
  year={2020},
  publisher={Springer}
}

@inproceedings{garg2020bae,
  title={BAE: BERT-based Adversarial Examples for Text Classification},
  author={Garg, Siddhant and Ramakrishnan, Goutham},
  booktitle={Proceedings of the 2020 Conference on Empirical Methods in Natural Language Processing (EMNLP)},
  pages={6174--6181},
  year={2020}
}

@inproceedings{yoo2021towards,
  title={Towards Improving Adversarial Training of NLP Models},
  author={Yoo, Jin Yong and Qi, Yanjun},
  booktitle={Findings of the Association for Computational Linguistics: EMNLP 2021},
  pages={945--956},
  year={2021}
}

@inproceedings{morris2020textattack,
  title={TextAttack: A Framework for Adversarial Attacks, Data Augmentation, and Adversarial Training in NLP},
  author={Morris, John and Lifland, Eli and Yoo, Jin Yong and Grigsby, Jake and Jin, Di and Qi, Yanjun},
  booktitle={Proceedings of the 2020 Conference on Empirical Methods in Natural Language Processing: System Demonstrations},
  pages={119--126},
  year={2020}
}

@article{zhang2015character,
  title={Character-level convolutional networks for text classification},
  author={Zhang, Xiang and Zhao, Junbo and LeCun, Yann},
  journal={Advances in neural information processing systems},
  volume={28},
  year={2015}
}

@inproceedings{Pang2005SeeingSE,
  title={Seeing Stars: Exploiting Class Relationships for Sentiment Categorization with Respect to Rating Scales},
  author={Bo Pang and Lillian Lee},
  booktitle={ACL},
  year={2005}
}

@inproceedings{Kim2014ConvolutionalNN,
  title={Convolutional Neural Networks for Sentence Classification},
  author={Yoon Kim},
  booktitle={EMNLP},
  year={2014}
}

@book{luke2009essentials,
author = { Sean Luke },
title = { Essentials of Metaheuristics},
edition = { second },
year = { 2013 },
publisher = { Lulu },
note = { Available for free at http://cs.gmu.edu/$\sim$sean/book/metaheuristics/ } }

@article{li2018textbugger,
  title={Textbugger: Generating adversarial text against real-world applications},
  author={Li, Jinfeng and Ji, Shouling and Du, Tianyu and Li, Bo and Wang, Ting},
  journal={arXiv preprint arXiv:1812.05271},
  year={2018}
}

@article{li2020contextualized,
  title={Contextualized perturbation for textual adversarial attack},
  author={Li, Dianqi and Zhang, Yizhe and Peng, Hao and Chen, Liqun and Brockett, Chris and Sun, Ming-Ting and Dolan, Bill},
  journal={arXiv preprint arXiv:2009.07502},
  year={2020}
}

@article{roth2021token,
  title={Token-Modification Adversarial Attacks for Natural Language Processing: A Survey},
  author={Roth, Tom and Gao, Yansong and Abuadbba, Alsharif and Nepal, Surya and Liu, Wei},
  journal={arXiv preprint arXiv:2103.00676},
  year={2021}
}

@article{guo2021gradient,
  title={Gradient-based Adversarial Attacks against Text Transformers},
  author={Guo, Chuan and Sablayrolles, Alexandre and J{\'e}gou, Herv{\'e} and Kiela, Douwe},
  journal={arXiv preprint arXiv:2104.13733},
  year={2021}
}

@electronic{artefact,
  url = {URL - TBC on publication},
  note = {URL - TBC on publication [Online; accessed 7-March-2025]},
  title = {Data and processing scripts for the paper ``Vulnerability of Natural Language Classifiers to
Evolutionary Generated Adversarial Text''},
  author = {Brownlee, Alexander E. I. and Singh, M. and },
  year = {2025}
}

@article{chelba2013one,
  title={One billion word benchmark for measuring progress in statistical language modeling},
  author={Chelba, Ciprian and Mikolov, Tomas and Schuster, Mike and Ge, Qi and Brants, Thorsten and Koehn, Phillipp and Robinson, Tony},
  journal={arXiv preprint arXiv:1312.3005},
  year={2013}
}

@article{moscato1992memetic,
  title={A memetic approach for the traveling salesman problem implementation of a computational ecology for combinatorial optimization on message-passing systems},
  author={Moscato, Pablo and Norman, Michael G},
  journal={Parallel computing and transputer applications},
  volume={1},
  pages={177--186},
  year={1992},
  publisher={Amsterdam.}
}

@article{nasar2021named,
  title={Named entity recognition and relation extraction: State-of-the-art},
  author={Nasar, Zara and Jaffry, Syed Waqar and Malik, Muhammad Kamran},
  journal={ACM Computing Surveys (CSUR)},
  volume={54},
  number={1},
  pages={1--39},
  year={2021},
  publisher={ACM New York, NY, USA}
}

\end{document}